\documentclass[runningheads]{llncs}
\usepackage{graphicx}
\usepackage{amsmath}

\begin{document}
\title{Whole slide image registration for the study of tumor heterogeneity 
}
%\thanks{Supported by organization x.}}
%
%\titlerunning{Abbreviated paper title}
% If the paper title is too long for the running head, you can set
% an abbreviated paper title here
%
%\author{Anonymous for review according to guidelines
%}
%\authorrunning{Anonymous}

\author{Leslie Solorzano\inst{1}
  \and
  B\'arbara Mesquita \inst{2,3}
  \and
  Gabriela M. Almeida \inst{2,3,4}
  \and
  Diana Martins \inst{2,3}
  \and
  Carla Oliveira \inst{2,3,4}
  \and
  Carolina W\"ahlby \inst{1}\thanks{European Research council for funding via ERC Consolidator grant 682810 to C. W\"ahlby.}
}
\authorrunning{L. Solorzano et al.}

% First names are abbreviated in the running head.
% If there are more than two authors, 'et al.' is used.
%

% \institute{
% Anonymous for review
% }

\institute{
Center for Image Analysis, Uppsala University \\
\email{\{leslie.solorzano, carolina.wahlby\}@it.uu.se}
\and
i3S, Instituto de Investiga\c c\~ao e Inova\c c\~ao em Sa\'ude, \\  
Universidade do Porto, Portugal
\and
Ipatimup, Institute of Molecular Pathology and Immunology, \\ University of Porto, Portugal
\and
Faculty of Medicine of the University of Porto, Portugal \\
\email{\{bmesquita, galmeida, dianam, carlaol\}@ipatimup.pt}
}

\maketitle  % typeset the header of the contribution

\begin{abstract}
Consecutive thin sections of tissue samples make it possible to study local variation in e.g. protein expression and tumor heterogeneity by staining for a new protein in each section. In order to compare and correlate patterns of different proteins, the images have to be registered with high accuracy. The problem we want to solve is registration of gigapixel whole slide images (WSI). This presents 3 challenges: (i) Images are very large; (ii) Thin sections result in artifacts that make global affine registration prone to very large local errors; (iii) Local affine registration is required to preserve correct tissue morphology (local size, shape and texture). In our approach we compare WSI registration based on automatic and manual feature selection on either the full image or natural sub-regions (as opposed to square tiles). Working with natural sub-regions, in an interactive tool makes it possible to exclude regions containing scientifically irrelevant information. We also present a new way to visualize local registration quality by a Registration Confidence Map (RCM). With this method, intra\nobreakdash-tumor heterogeneity and charateristics of the tumor microenvironment can be observed and quantified.

\end{abstract}

\section{Introduction}

In digital histopathology, whole slide imaging (WSI), denotes the scanning of standard glass slides containing tissue to produce digital images that can be stored, analyzed, annotated and shared remotely and locally~\cite{dig_patho,ameisen}. With WSI comes a set of challenges, comparable to geographical information systems and astronomical images. WSI have sizes ranging from 35,000 to 200,000 pixels\textsuperscript{2} . These images require special formats, readers, standards and hardware to be read and processed. Processing of this type of images requires either computers with enough memory or smart strategies for partitioning the data without losing the spatial context. 

\begin{figure}[t]
    \centering
    \includegraphics[width=\columnwidth]{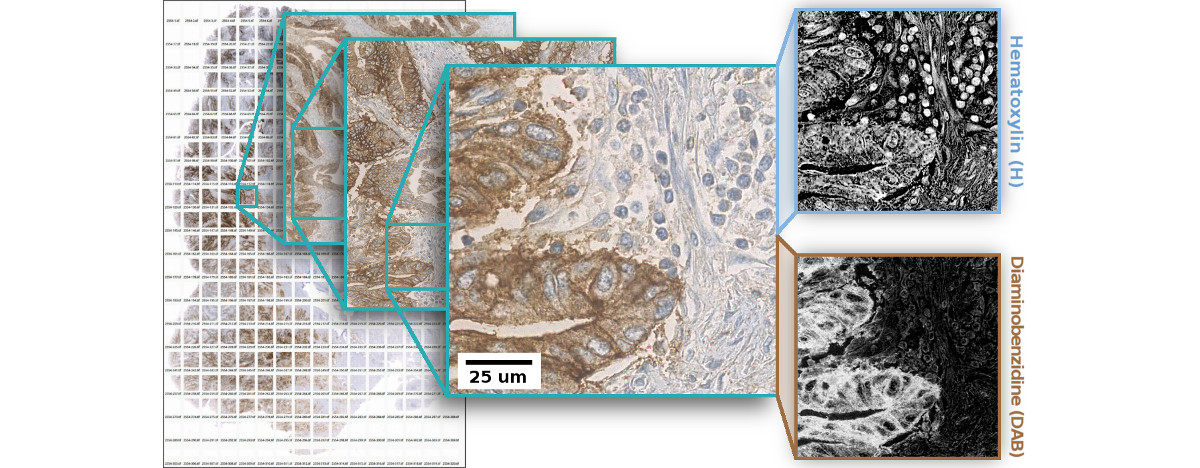}
    \caption{\textit{Left) 320 5000 pixels\textsuperscript{2} tiles are represented. One tile is ``zoomed in'' up to a full resolution of 0.25 micrometers per pixel. Right) Two colors for H and DAB are shown in separate tiles after color deconvolution.} }
    \label{fig:resolutionsDeconv}
\end{figure}

Usually, a tissue section scanned with brightfield microscopy is stained with two or more stains. The tissue is "photographed" with visible light and the result is a microphotograph collected in 8-bit pixels in RGB space. Figure~\ref{fig:resolutionsDeconv} shows an example of a WSI of gastric cancer tissue stained with Hematoxylin and Diaminobenzidine (H and DAB). H stains for sub\nobreakdash-cellular compartments present in all the tissue such as the cellular membranes, cytoplasms and nuclei, allowing to delimit and resolve objects of interest. DAB on the other hand stains for a specific protein only present in certain cell types and sub\nobreakdash-cellular compartments.

Conventional image processing libraries are not designed to process WSI. Most of them were conceived to index up to $2^{32}$ pixels and fail when naively prompted to open special headers and formats like those used for WSI. Images can be stored in a pyramidal fashion, where several resolutions of the image are saved in the same file (or in separate files) and can be accessed according to the desired resolution level. In order to create correct and efficient tools for pathologists, new and faster methods for image processing, such as convolution, filtering, registration, annotation and visualization have to be developed for this special type of images. 

In the study of tumor microenvironments (TME) and intra\nobreakdash-tumor heterogeneity, consecutive thin sections of tissue are stained so that a common structure is revealed ~\cite{spagnolo,xavier}, for instance with H, while each section is stained separately with DAB for a different protein. These sections are then scanned into a WSI that can be partitioned into new image channels by color deconvolution ~\cite{ruifrok}, making it possible to process general tissue information from H, and specific information from DAB, separately. After the jump from analog to digital pathology, %there is an increasing need for methods 
new methods are needed to help automate the analysis of big amounts of WSI. But each WSI is not useful on its own, information has to be spatially aligned and there is a need for better methods of WSI registration and assessment of its quality.

% \begin{figure}
%     \centering
%     \includegraphics[width=\columnwidth]{images/3dstruct-l.jpg}
%     \caption{\textit{Spatial localization of the same glands (1 to 3) in three consecutive sections (A,B and C) of the same tissue, plus a 3D illustration of the glands. The same structure will appear smaller or disappear between sections.}  }
%     \label{fig:3dstruct}
% \end{figure}

Pixel perfect alignment in WSI is impossible, particularly because structures in consecutive tissue sections should appear in different locations and sizes. This is a common problem in 3D tissue reconstruction. %Figure ~\ref{fig:3dstruct} shows three locations that belong to the same glands but in different sections. 
A choice has to be made, either the image is deformed to try and achieve pixel perfect alignment or a different transformation is chosen. Image registration (image spatial alignment) is commonly expressed as an optimization problem which can be solved by iteratively searching for transformation parameters such that the distance between a pair of points (feature-based) or images (intensity-based) is minimized. The transformation can be rigid, affine or non\nobreakdash-rigid (deformable, or non\nobreakdash-linear). An affine transformation is done to preserve points, parallel lines and ratios between structures. 

In research, many kinds of registration methods are continuously experimented with, but according to ~\cite{medimreg2016}, lack of genericity and robustness has prevented the inclusion of deformable methods in commercial software for clinical diagnoses while globally rigid registration is the most frequently used approach. The reason is the ability to control and study the parameters for registration. With affine methods, fewer parameters have to be set as opposed to deformable approaches, and the distance between spatial landmarks can be minimized without compromising the structural integrity of the tissue.  In affine transformations, the values of the affine matrix represent the amount of stretching and skewing which can be constrained. %maybe%%% Two types of image registration are commonly used, pixel intensity based, and feature based. %%maybe remove the prev line later
In multimodal imaging, entropy based measures are the most common distance measure in intensity based registration~\cite{histoImAna09}. Within the feature based registration approaches, a spatial correlation between features is used to find the transformation. %address reviewer 1 
Depending on the desired resolution the alignment is expected to achieve, different registration schemes can be planned, where successive 
registration methods can be applied. There is no silver bullet and some methods may become irrelevant for different types of applications. It is very common to start with a rougher and less expensive global alignment in lower resolutions, followed by affine or non\nobreakdash-rigid deformations. For instance, in histology images, \cite{xavier} uses a global rigid transformation followed by local non\nobreakdash-rigid transformations of selected ROIs guided by intensity based information. In \cite{cooper,trahearn} there are: a feature based rigid transformation followed by feature based local refinements. Compared to previous works, our applications require cellular alignment, and a quantifiable confidence for it.

The main contributions of this paper are a piece-wise approach for image registration and visualization of registration confidence. The piece-wise approach lets us overcome the challenges of handling large WSI and tissue artifacts.  We create a Registration Confidence Map (RCM) as well as a stain Combination Quantication Map (CQM) to visualize local structure and colocalization of proteins within the tissue sample. %address reviewer 1 
Automated and semi-automated registration methods require fast and efficient visualization of the images to immediately detect or correct mistakes. Even when handling big amounts of data, visualization is important before making any kind of prognostic assessment.

\section{Materials and Methods}

\subsection{Image data}

We received a set of WSI of thin consecutive sections of gastric tumor, where each section spans the thickness of a single cell. All sections were exposed to the same H, but each DAB represents a unique protein per tissue section. Three proteins codenamed A, B and C are presented here. Pixel size is 0.25$\mu$m, and the full images are in average 90000 pixels\textsuperscript{2}. Despite careful sample preparation, artifacts such as folds, rips and loss of tissue were observed in all images. 

\subsection{WSI preprocessing}

We first separate H which reveals common tissue structures from the specific DAB in all slides. In the end, two images of the same size as the original are saved, one representing H and one representing DAB. For this purpose, we calculated a stain matrix so that a pixel in RGB space can be projected to an H and DAB space. We acquired a new palette on each tissue section. We implemented color deconvolution based on the work of ~\cite{ruifrok,wemmert}, the optical density (OD) matrix is calculated and then its inverse allows to send RGB points to H and DAB space. 

After color deconvolution we contrast stretched all channels to the 99th percentile followed by a gamma correction with gamma 1.85. This is important %maybe%%% due to the fact that 
since even the same stain can produce different results depending on how much it can actually penetrate a cell and parts of the tissue. It is important to note as well that in this case we are not quantifying the amount of a given stain but its presence. There has been much controversy as to decide if the pixel intensity represents the amount of stain and the nature of this representation ~\cite{vanderlaak}.

\begin{figure}[t] 
    \centering
    \includegraphics[width=\columnwidth]{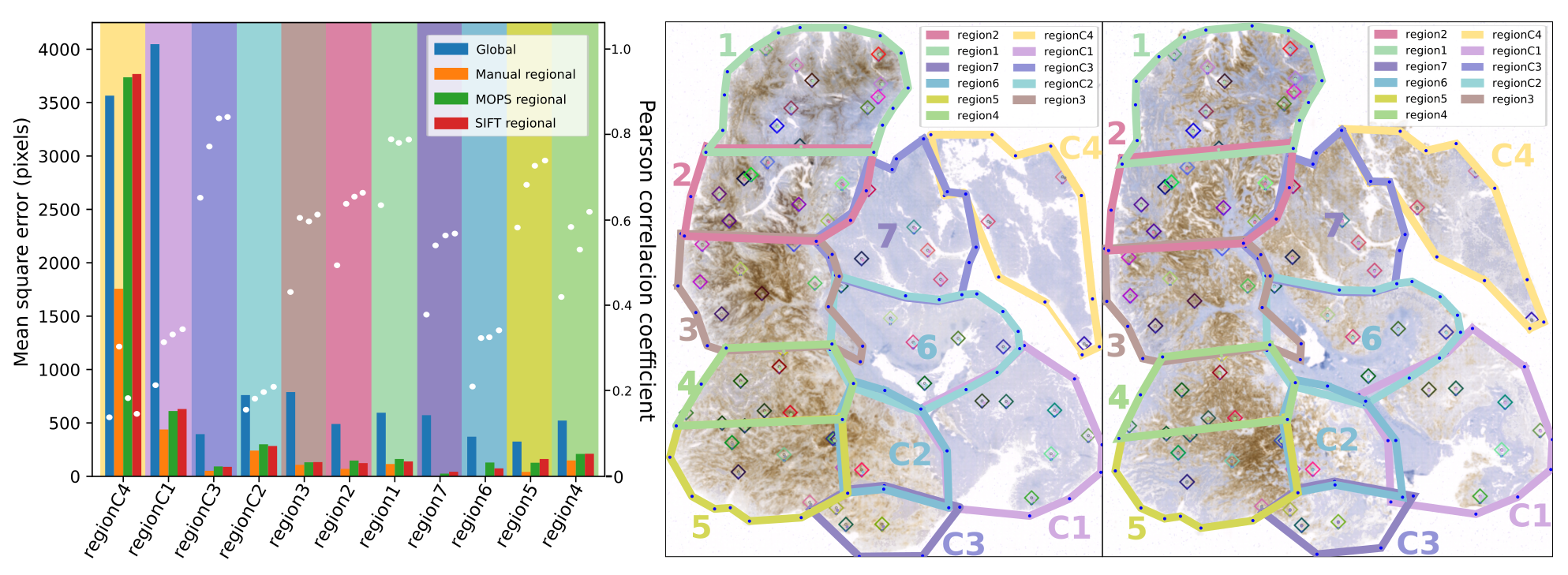}
    \caption{\textit{Regions, MSE of landmarks after registration and PCC for each region. Bars show the MSE of distance between landmarks in pixels and the white dots show the PCC where 1 means complete colocalization.  }  }
    \label{fig:MSE}
\end{figure}

As part of the WSI preprocessing we also developed a web-interface for manual selection of pairs of natural sub\nobreakdash-regions and control points. The web interface is based on OpenSeaDragon ~\cite{openseadragon} and enables quick zooming, panning and selection of regions. Image artifacts such as folds and tears are easy to spot visually, and we selected pairs of high-quality image regions from consecutive tissue sections. An example of sub regions is shown in Figure~\ref{fig:MSE}. These sub\nobreakdash-regions served as input for feature selection and registration as described below.

\subsection{Feature extraction and image registration}

After regions are selected, common spatial features were found in each pair of corresponding regions and then these features were matched thus finding a transformation. Our innovation comes from the use of only the H channel, the channel that contains the common tissue structure and morphology.

To extract features we compare three methods, namely Scale Invariant Feature Transform (SIFT), Multi\nobreakdash-Scale Oriented Patches (MOPS) and manually using a visualization and tool we developed for this purpose. Using Fiji ~\cite{imagej_sift}, a list of possible pairs of feature sets are found and matched using RANSAC.

We thereafter found a transformation by matching features using the Blendenpik~\cite{blendenpik} least squares minimization as used by Numpy. This results in a transformation matrix that is applied either to a specific region or to the whole tissue. If regions are used, one must ensure that the regions have sufficient overlap to avoid introducing holes in the resulting image in case a transformation becomes too big. Nevertheless big transformations are not expected between adjacent regions.

\subsection{Evaluation and creation of RCM and CQM}

We evaluated global registration results using the Pearson correlation coefficient (PCC) ~\cite{oheim} with the two H as input, including all pixel pairs where at least one of the pixels was greater than zero. Next, we also evaluated the image registration result by defining colocalization as the percentage of pixel pairs where both intensity values are above the Costes threshold ~\cite{costes} which estimates the maximum threshold for each color below which pixels do not show any statistical correlation. Table ~\ref{table:pearson_table} presents the PCC and the percentages of colocalized image pixels for each of the proposed approaches.

\begin{table}
\centering
\caption{PCC table for proteins A and B. }\label{table:pearson_table}
\begin{tabular}{|c|c|c|c|c|c|c|}
\hline
Images	& PCC total	& PCC coloc	& \%A Vol	& \%B Vol	& \%A \textgreater th	& \%B \textgreater th\\
\hline
Affine global	& 0.438	& 0.1437	& 63.32\%	& 71.45\%	& 66.03\%	& 72.95\%\\
Manual regional	& 0.697	& 0.6387	& 85.36\%	& 86.81\%	& 88.50\%	& 89.35\%\\
MOPS regional	& 0.696	& 0.6417	& 85.10\%	& 86.60\%	& 88.46\%	& 88.98\%\\
SIFT regional & 0.697	& 0.6475	& 85.10\%	& 86.58\%	& 88.47\%	& 88.97\%\\
\hline
\end{tabular}
\end{table}

%PCC offers an effective quantitative way to calculate colocalization in image analysis since it can provide one single value. It ranges from -1 to 1 which might seem counter intuitive but it is effective as long as the number of compared pixels is more or less the same, which is the current case since the common H stain represents approximately the same amount of tissue. 

Next we created the Registration Confidence Map (RCM) by subsampling the WSI by a factor 200. Using the Costes thresholds described above, output pixels were either colorcoded white (A\textgreater t and B\textgreater t), red (A\textgreater T, B $\leq$ t), green (A$\leq$t, and B\textgreater T) or black (A$\leq$t, B$\leq$t). 

Finally, a map of protein colocalizations was created, referred to as the Colocalization Quantification Map (CQM). Again, Costes thresholds were applied and each pixel color coded based on what combination of stains reached above the intensity threshold. An example is shown in in figure~\ref{fig:CQM}.

\begin{figure}[t]
    \centering
    \includegraphics[width=\columnwidth]{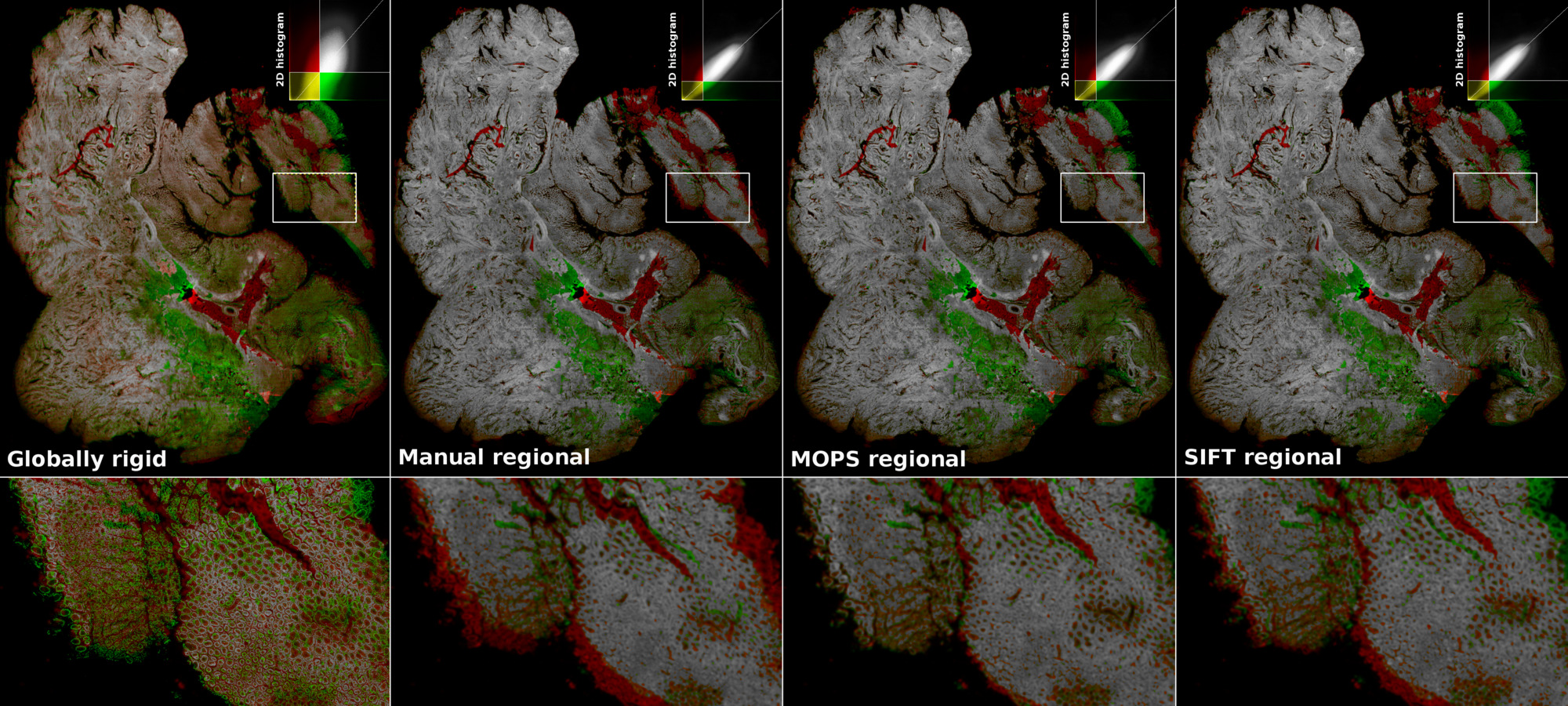}
    \caption{\textit{RCM of stain A and B. Global affine, manual regional, MOPS regional and SIFT regional approaches to registration. Each has it's own 2D histogram color coded with the frequencies of intensity pairs. Thresholds are also shown. White means the pixels are colocated, while red and green shows regions where the H staining for slide A or B do not match. A closeup within region C4 is added to show a higher level of detail.}  }
    \label{fig:RCM}
\end{figure}
\section{Results}

\begin{figure}[t]
    \centering
    \includegraphics[width=\columnwidth]{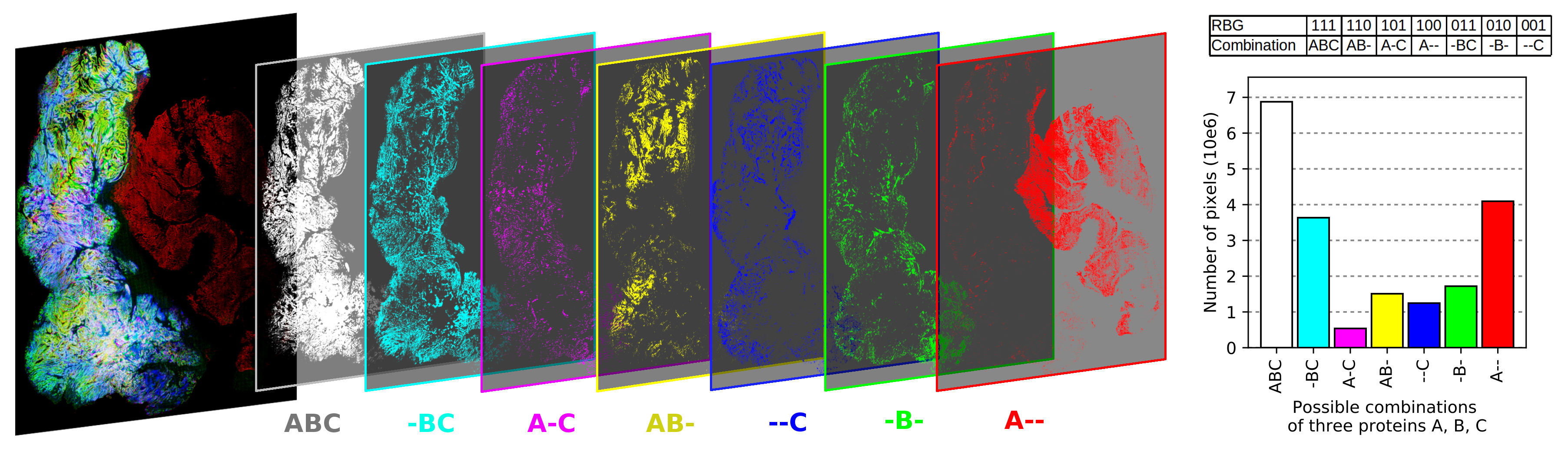}
    \caption{\textit{CQM of three slides stained for different proteins A, B and C.  Different combinations of proteins can be analyzed and their colocalization quantified per pixel. Given that three proteins are being studied, the results are shown in RGB color space. The plot shows the amount per protein combinations in the WSI.  %Localized quantification of these combinations is of interest to study of TME and ITH.  %Each color indicates presence/non-presence per pixelPair-wise colocalization may be displayed as a heat-map, but as soon as more than two proteins are compared, a binary representation turns out to be more informative. 
    }  }
    \label{fig:CQM}
\end{figure}

In the presented results, three proteins codenamed A, B and C are registered and analyzed for colocalization in the tissue. Their CQM and RCM are shown in figure~\ref{fig:RCM} and~\ref{fig:CQM}, respectively.

Image registration can be evaluated by finding distance between landmarks after transformation, using intensity based methods, or by comparison of transformation parameters %(more common in non\nobreakdash-linear or deformable registration). 
As seen in ~\cite{curt}, we consider landmark and contour based methods insufficient for our purpose. Distance between landmarks can be minimized but this does not guarantee alignment of the rest of the tissue. For this reason we consider RCM a good measure to find the locations within the tissue sections that can be trusted and to which degree. Nevertheless in figure~\ref{fig:MSE} we offer a comparison of the mean square error (MSE) of landmark distances for each %transformation 
approach. It is worth noting that in region C4 points have been manually selected intentionally to force a fit of the tissue. For this reason the manual landmarks approach optimizes the distance between landmarks and shows an apparently good result while it can be visually assessed in figure~\ref{fig:RCM} that region C4 is not properly aligned and that probably should be separated in several regions. Additionally, regions C2 and C3 each achieve a small distance between landmarks yet they have very different PCC. This confirms that landmark based evaluation should not be the only method to evaluate registration. In table~\ref{table:pearson_table} several values are offered to quantify and interpret what is observed in the RCM.

\begin{itemize}
\item PCC total: PCC for all the non zero\nobreakdash-zero pixels in the image
\item PCC coloc: PCC for pixels where both H levels are above their respective threshold (white box in the 2D histogram)

\item \%A vol (and \%B vol): the number of pixels in the white box divided by the sum of white and green (or red for B) boxes.

\item \%A \textgreater t (and \%B \textgreater t): the sum of the intensities of the pixels in the white box divided by the sum of intensities on both white box and green box (or red for B).
%\item \%A vol and \%B vol : \%A is the number of pixels in the white box divided by the sum of white and green boxes. \%B is the number of pixels in the white box divided by the sum of white box and red box
%\item \%A \textgreater t and \%B \textgreater t:  For A is the sum of the intensities of the pixels in the white box divided by the sum of intensities on both white box and green box. For B is the sum of the intensities of the pixels in the white box divided by the sum of intensities on both white box and red box.
\end{itemize}

Figure~\ref{fig:CQM} shows the overlap in expression of three proteins, all registered using our method. These CQM will be used in overlap quantification for tumor characterization studies.

\section{Conclusions}

We present a new way to do WSI registration and a new way to evaluate it based on the comparison of tissue morphology by using PCC as well as spatial maps of RCM and CQM. This general framework could be used for any staining protocol (HDAB, H\&E, etc) so long as the stain used for matching is comparably consistent across tissue sections. These maps do not only provide an excellent visual representation of spatial heterogeneity of tumor tissues, but can also serve as computer-generated input for training deep convolutional neural networks. Many learning approaches today rely on manual annotations, which are often highly variable between expert pathologists. Tumor niches and normal tissue can instead be automatically defined from our CQM, and Deep convolutional neural networks (DCNNs) can be trained to detect corresponding structural information (from the H-channel) in a clinical setting. This tool is expected to improve our understanding on ITH with a potential impact for definition of personalized therapies.

% ---- Bibliography ----
%
% BibTeX users should specify bibliography style 'splncs04'.
% References will then be sorted and formatted in the correct style.
%
\bibliographystyle{splncs04}
% \bibliography{mybibliography}
%

\end{document}